\documentclass[letterpaper, 10 pt, journal, twoside]{ieeetran}

\IEEEoverridecommandlockouts                              

\usepackage{geometry}
\usepackage[T1]{fontenc} 

\usepackage{amsmath,amsfonts}
\usepackage{algorithmic}
\usepackage{array}
\usepackage{textcomp}
\usepackage{stfloats}
\usepackage{url}
\usepackage{verbatim}
\usepackage{graphicx}
\def\BibTeX{{\rm B\kern-.05em{\sc i\kern-.025em b}\kern-.08em
    T\kern-.1667em\lower.7ex\hbox{E}\kern-.125emX}}
\usepackage{balance}

\usepackage[caption=false, font=footnotesize]{subfig}
\usepackage[footnotesize,bf]{caption}
\usepackage{graphicx,psfrag,amsmath,amsfonts,verbatim,url,booktabs,array,float}
\usepackage[short]{optidef}
\usepackage{censor}
\usepackage{graphicx}
\usepackage{bm}
\usepackage{subfloat}
\usepackage{sidecap}
\usepackage[hidelinks]{hyperref}

\usepackage{xcolor}
\definecolor{orange}{RGB}{255,127,0}
\definecolor{cyan}{RGB}{0,255,255}
\definecolor{magenta}{RGB}{255,0,255}

\newcommand{\R}{\mathbb{R}}

\definecolor{nicegreen}{rgb}{0.1, 0.6, 0.2}

\renewcommand{\baselinestretch}{0.987}

\usepackage{pgfplots}
\pgfplotsset{compat=1.16}
\usepackage{tikz}
\usepackage{tikzscale}
\usetikzlibrary{matrix}
\usetikzlibrary[pgfplots.groupplots]
\usetikzlibrary{arrows.meta}
\usetikzlibrary{backgrounds}
\usepackage{pgfplots}
\usepgfplotslibrary{groupplots}

\begin{document}

\title{
Differentiable Physics Simulation of\\ 
Dynamics-Augmented Neural Objects
}

\author{Simon Le Cleac'h$^{1,3}$, Hong-Xing Yu$^{2}$, Michelle Guo$^{2}$, Taylor Howell$^{3}$, Ruohan Gao$^{2}$, \\Jiajun Wu$^{2}$, Zachary Manchester$^{3}$ and Mac Schwager$^{1}$%
\vspace{-4mm}
\thanks{Toyota Research Institute (TRI) provided funds to support this work. This work is also in part supported by Qualcomm, Amazon, NSF CCRI \#2120095, NSF RI \#2211258, ONR MURI N00014-22-1-2740, and the Stanford Institute for Human-Centered AI (HAI). 
\it{(Corresponding author: Simon Le Cleac'h. {\tt\footnotesize simonlc@stanford.edu}) }} 
\thanks{$^{1} $Multi-robot Systems Laboratory, Stanford University, California, USA.
        }%
\thanks{$^{2} $Stanford Vision Laboratory, Stanford University, California, USA.
        }%
\thanks{$^{3} $Robotic Exploration Laboratory, Carnegie Mellon University, Pennsylvania, USA.
        }%
}

\maketitle

\begin{abstract}
    We present a differentiable pipeline for simulating the motion of objects that represent their geometry as a continuous density field parameterized as a deep network. This includes Neural Radiance Fields (NeRFs), and other related models. From the density field, we estimate the dynamical properties of the object, including its mass, center of mass, and inertia matrix. We then introduce a differentiable contact model based on the density field for computing normal and friction forces resulting from collisions. This allows a robot to autonomously build object models that are visually and \emph{dynamically} accurate from still images and videos of objects in motion. The resulting Dynamics-Augmented Neural Objects (DANOs) are simulated with an existing differentiable simulation engine, Dojo, interacting with other standard simulation objects, such as spheres, planes, and robots specified as URDFs. A robot can use this simulation to optimize grasps and manipulation trajectories of neural objects, or to improve the neural object models through gradient-based real-to-simulation transfer. We demonstrate the pipeline to learn the coefficient of friction of a bar of soap from a real video of the soap sliding on a table. We also learn the coefficient of friction and mass of a Stanford bunny through interactions with a Panda robot arm from synthetic data, and we optimize trajectories in simulation for the Panda arm to push the bunny to a goal location. \href{https://youtu.be/Md0PM-wv_Xg}{Video: $\texttt{youtu.be/Md0PM-wv\_Xg}$}
\end{abstract}


\section{Introduction}
    We present the Dynamics-Augmented Neural Object (DANO), a novel object representation that augments a neural object with dynamical properties, so that its motion under applied forces and torques can be simulated with a differentiable physics engine. We also propose a method for computing contact forces due to collisions between the DANO and other objects in the simulation.  We specifically focus on neural objects that are trained from RGB images only, and encode their geometry through a neural density field---a continuous density field represented as a deep neural network, such as the Object-centric Neural Scattering Function (OSF \cite{yu2023learning}), which is an object-centric NeRF \cite{mildenhall2020nerf,yang2021learning}, or related models \cite{gao2021ObjectFolder,gao2022ObjectFolderV2}. 
    DANOs capture both geometric and dynamical properties of rigid objects from the underlying neural density field.
    This allows a robot to build a dynamical simulation of an object from observed RGB images, and fine-tune that model through real-to-sim transfer using videos of the object in motion. 

    \begin{figure}[t]
    	\begin{center}
    	    \includegraphics[width=1.0\columnwidth]{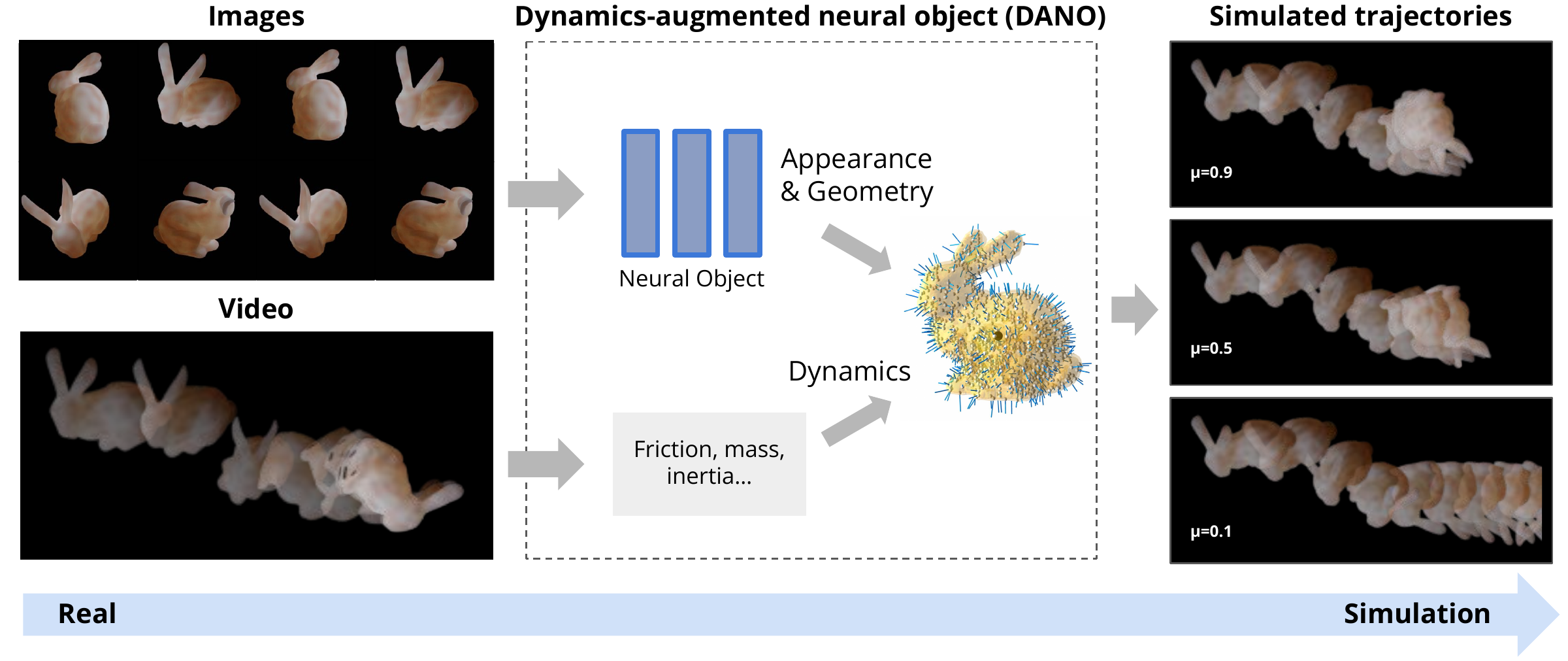}
       	\end{center}
        \vspace{-2mm}
    	\caption{Our pipeline for simulating the motion of neural objects. A neural object model such as a NeRF is trained from still images (top left), and object trajectories are extracted from videos of the object in motion (bottom left). Object mass and friction properties are computed and candidate surface points and normals are sampled to produce a Dynamics-Augmented Neural Object (DANO) (middle). The DANO is simulated in a differentiable physics simulator interacting with planes, robots, and other rigid objects (right). The resulting simulation can be used for real-to-simulation transfer, or to synthesize robot behaviors, e.g., for manipulation.}
    	\label{pipeline}
    	\vspace{-6mm}
    \end{figure}

The main challenge with neural density fields is that they do not give a distinct object surface, and are better interpreted as a differential probability of occupancy.  Since contact forces (friction and normal forces) arise from interactions between surfaces, how does one simulate contact for objects with neural density fields, which have no distinct surface?  One naive approach is to choose a single level set of the neural density to stand in as the object surface, then compute a traditional mesh model from this level set using, e.g., marching cubes \cite{lorensen1987marching} and simulate motion with the resulting mesh.  We show that this leads to poor-quality meshes with significant spurious artifacts that prevent accurate simulation.  Instead, we propose an inherently probabilistic differentiable model of contact for objects with neural density fields, and derive Monte Carlo techniques for computing contact forces under this model.  We show that this model can give high-fidelity simulation trajectories nearly indistinguishable from trajectories simulated with ground-truth knowledge of the 3D object geometry.

\begin{figure}[t]
    	\begin{center}
    	\includegraphics[height=0.22\textwidth]{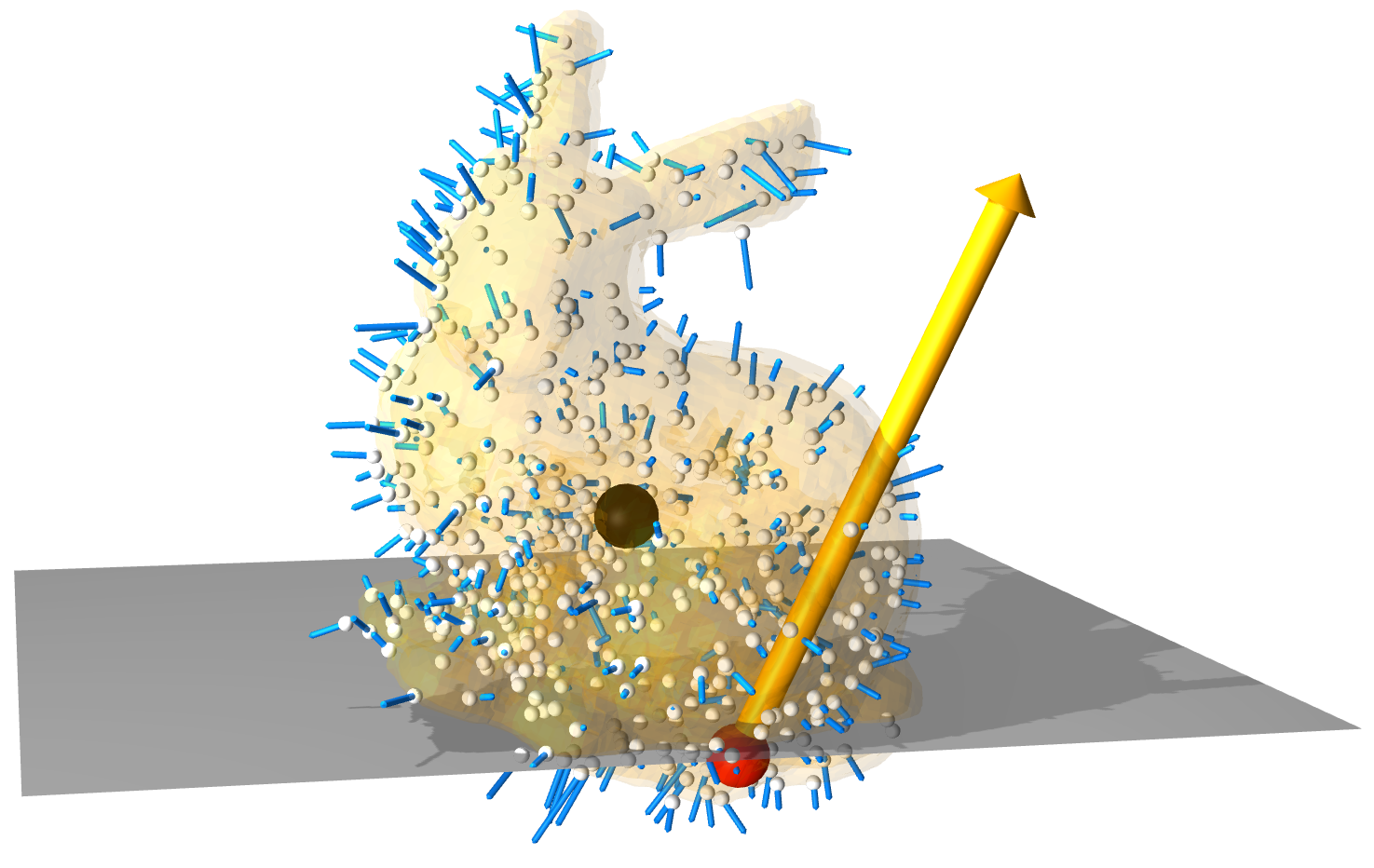}
    	\end{center}
        \vspace{-5mm}
    	\caption{DANO obtained from a neural object model of the Stanford bunny. We  sample points (white dots) from the neural density field and compute the field gradients (blue arrows) to obtain an approximate outward normal, which is used to compute contact forces in our simulator. We estimate the mass, center of mass (black dot), and inertia matrix by integrating over the neural density field.
    	A contact force (yellow arrow) due to the collision between the DANO and the plane is applied at the centroid of the overlap volume (red dot), which is exaggerated for visualization.  
    	}
        \vspace{-5mm}
    	\label{fig:preprocess}
    \end{figure}

Our method is illustrated in Fig.~\ref{pipeline}. We first train a neural object model to represent the object geometry implicitly as a neural density field.  We then compute a Monte Carlo estimate of the mass, center of mass, and inertia matrix from the neural density field, up to an unknown mass scale factor\footnote{The mass and inertia matrix can only be found up to a scale factor from images and video alone. The mass scale factor can be resolved by observing object trajectories with known forces, or contact interactions with other objects of known mass, such as a known robot arm.}. We propose a Monte Carlo approach to simulate the contact forces (normals and friction forces) based on sampled candidate surface points and normals, as shown in Fig.~\ref{fig:preprocess}, and integrate this contact model within an existing differentiable physics simulator, Dojo \cite{howelllecleach2022}.  Using DANO models in Dojo, we demonstrate real-to-sim transfer by obtaining, from real video sequences of a sliding bar of soap, an estimate of the coefficient of friction of the soap. We also estimate, from synthetic trajectory data, the coefficient of friction and the missing mass scale factor for a neural object model of a Stanford bunny. Finally, we leverage the differentiability of the simulation pipeline to optimize a trajectory in simulation for a robot arm to push the Stanford bunny to a goal location.
Our key contributions are:

    \begin{itemize}
        \item Estimating inertial and friction properties and sampling candidate surface points and normals from a neural density field to form a Dynamics-Augmented Neural Object (DANO).
        \item A differentiable formulation of rigid-body contact forces for DANOs.
        \item Integration of DANO models with the Dojo differentiable physics engine. 
        \item Demonstration of the differentiable pipeline to learn physical properties from real videos and synthetic data, and to optimize a pushing trajectory for a robot arm.
    \end{itemize}

    \section{Related Work}
    Extracting accurate object models from real perception data to allow a robot to plan and interact with those objects is a grand challenge in robotics with substantial existing work from multiple related fields.
    
    \textbf{Dynamic NeRF.}
    Recent years have witnessed an explosion in neural scene representations~\cite{sitzmann2019scene,park2019deepsdf,mescheder2019occupancy}. In particular, Neural Radiance Fields (NeRF)~\cite{mildenhall2020nerf} have shown impressive novel view synthesis results from real-world images without any 3D input, indicating the promise to reconstruct both appearance and geometry from only RGB images. To allow modeling dynamics in addition to appearance, there have been various extensions to NeRF for deformable objects~\cite{park2021nerfies,pumarola2021d,tretschk2021non,park2021hypernerf,liu2021neural} and general motions~\cite{li2021neural,xian2021space,du2021neural,peng2021neural,gao2021dynamic}. However, these methods only fit given dynamic events from videos without generalization ability to new scenes or novel motions. The closest to our work are a few recent works that aim to learn dynamic NeRFs for planning and control~\cite{driess2022learning,li20223d,shen2022acid}. Nevertheless, all of these focus on learning to approximate dynamics by black-box latent representations without accounting for the underlying physics, and thus they cannot generalize to unseen motions or scenarios. In contrast, our approach explicitly infers dynamic physical parameters, and simulates motion in a dynamics engine that respects the known laws of Newtonian mechanics.  Because we rely on a physics engine instead of a learned dynamics model, we can predict motions in scenarios that are arbitrarily different from the training data, including previously unseen collisions between multiple neural objects, or between a neural object and URDF robot models, or arbitrary arrangements of shape primitives such as half-spaces, spheres, and capsules.
    
    \textbf{Real-to-Simulation.}
    Some previous works have relied on videos of the object in motion to identify parameters of a dynamical model \cite{le2021differentiable, shen2022acid,sundaresan2022diffcloud, pfrommer2020contactnets,murthy2020gradsim}. ACID~\cite{shen2022acid} and DiffCloud~\cite{sundaresan2022diffcloud} focus on identifying models for \emph{deformable} objects, while we focus on rigid objects in this paper. ContactNets~\cite{pfrommer2020contactnets} regresses a rigid contact model from observed 3D object trajectories obtained from videos of objects with AprilTags \cite{olson2011apriltag}.  ContactNets starts with a geometrical object model, which can be fine-tuned through this trajectory data.  In contrast, we obtain our DANO model from untagged RGB images without prior geometrical information, and fine-tune mass and friction coefficients through untagged videos. Finally, GradSim~\cite{murthy2020gradsim} combines a differentiable rendering tool and a differentiable physics simulator. Our work differs from GradSim in two respects: GradSim represents objects as meshes, while we leverage the neural density field learned directly from images; GradSim demonstrates impressive system identification capabilities in simulated environments, while we further perform experiments with real-world data.

    \textbf{Differentiable Simulation.} Differentiable simulators promise the ability to back-propagate gradients through a simulation rollout, to allow for optimization of simulation parameters to fit observed data (System Identification), or to optimize robot motions or robot policies. Several differentiable rigid-body simulators have been recently proposed \cite{howelllecleach2022, de2018end, werling2021fast, freeman2021brax, heiden2021neuralsim}. They handle contact for simple shape primitives (spheres, capsules, etc.) and more complex models decomposed into a union of convex shapes. However, none of them support the simulation of neural objects represented by density fields. In this work, we propose a novel differentiable contact model for rigid bodies represented by a neural density field. We seamlessly embed this contact model into an existing differentiable simulator, Dojo~\cite{howelllecleach2022}. Note that the proposed DANO and contact model can also be integrated with other implicit integration-based differentiable simulators~\cite{werling2021fast, heiden2021neuralsim}. 
    
    \section{Object-Centric Neural Scattering Function (OSF) Model}
    \label{sec:technical_background}
    We use a specific object-centric model\footnote{Object-centric means that the model represents a single object in a body-fixed frame, and can be reposed through a rigid body transform with respect to a global frame.  In contrast, the large majority of neural scene representations (like NeRF) have no notion of individual objects or poses.} called the Object-Centric Neural Scattering Function (OSF)~\cite{yu2023learning}.   Unlike NeRFs, which assume a static scene with fixed illumination, OSFs are relightable and compositional. Thus, OSFs allow representing dynamic scenes where objects can move and change appearances (e.g., an object's shadow changes with its pose), while NeRFs do not support this purpose.  OSFs are also relightable, meaning they can render from a variety of different lighting conditions. 
    
    More specifically, an OSF models a volumetric function $(x, \omega_\text{light}, \omega_\text{out}) \rightarrow (\phi, \rho)$, where $\rho = (\rho_r, \rho_g, \rho_b) \in \mathbb{R}^3$ is the cumulative radiance transfer to model the object appearance and $\phi(x)$ is the volumetric density that models the object geometry. The $\rho$ function takes as input a 3D point $x \in \mathbb{R}^3$ in the object coordinate frame, an incoming distant light direction  $\omega_\text{light}$, and an outgoing radiance direction $\omega_\text{out}$ at that location, while the density $\phi(x)$ only requires the 3D point. The OSF can be re-posed in a background scene, or with respect to other neural objects through a pose transform applied to $x$, $\phi(Rx+\tau)$, where $(R, \tau)$ are a rotation matrix and translation vector, respectively, defining the pose of the object.  In this work, we use the learned density field $\phi(x)$ of the object as a proxy for modeling its geometry. We do not use the appearance information $\rho$, though inferring dynamical properties such as mass density and friction coefficient from appearance is a promising direction for future work.

    \section{Dynamics Augmented Neural Objects (DANOs)}
        \label{sec:dano}
    In this section, we describe how to obtain a DANO from a neural density field by estimating the mass, center of mass, and inertia matrix directly from the field. We formulate a probabilistic contact model, and detail computations for contact forces between the DANO and a shape primitive (e.g. a half-space, sphere, or mesh), and between two DANOs.  Ultimately, we integrate this contact model into the Dojo simulator \cite{howelllecleach2022}, which uses the implicit function theorem to differentiate through the integrator\footnote{
        For tractability reasons, the gradient through collision detection is approximate, e.g. it does not take into account the differentiation of the contact normal between two neural objects.
        }.
        
    \textbf{Estimating Mass, Center of Mass, and Inertia Matrix.} In an offline phase, we estimate the object's mass, inertia matrix, and center of mass by treating the neural density field $\phi(x)$ as a scaled mass density field.  We compute Monte Carlo estimates of the integrals that define the mass, inertia matrix, and center of mass in terms of the mass density.  Specifically, we uniformly sample a set of points $X$ from the density field $\phi(x)$, and compute the mass, inertia matrix, and center of mass as:
        \vspace{-3mm}
        \begin{align}
            \bar{\phi}(x) = &\alpha \phi(x), \\
            \label{eq:MassScale}
            m = &\int_{x \in \R^3} \bar{\phi}(x) dx \approx \sum_{x \in X} \bar{\phi}(x), 
        \end{align}
        \begin{align}
            \mu = &\frac{1}{m} \int_{x \in \R^3} x \bar{\phi}(x) dx \approx \frac{1}{m} \sum_{x \in X} x \bar{\phi}(x),\\
            J = &\int_{x \in \R^3} \bar{\phi}(x) (x^Tx I - x x^T) dx \approx \nonumber \\
            &  \sum_{x \in X} \bar{\phi}(x) (x^T x I - x x^T),
        \end{align}
        where we denote $\bar{\phi}$ the volumetric mass density field, $\alpha$ the scaling factor, $m$ the mass, $J$ the moment of inertia, and $\mu$ the center of mass. The parameter $\alpha$ is the unknown mass scale factor that can only be estimated from observed interactions between the object and known forces or other objects of known mass.  We discuss the identification of this parameter, along with friction parameters, in Sec.~\ref{sec:applications}.
        
        We note that the scaled density field $\bar{\phi}$ does not align well with the full mass distribution of the real object on a point-by-point basis. However, as illustrated in the experiments in Sec.~\ref{sec:applications}, we find empirically that integrating over the density field does give close enough approximations to the mass, center of mass, and moment of inertia matrix to provide dynamically plausible simulations. 
        These estimates can serve as initial guesses for a system identification method to further refine them to match the real mass, center of mass, and inertia matrix of the object. 
        The scaled density field $\bar{\phi}$ is the best guess we can make from still images. However, incorporating information from a video of the object in motion would allow for refinement of these initial estimates and leads to more a accurate estimation of these quantities.
    
    \textbf{Probabilistic Contact Model.} 
        Given the mass, centroid, and inertia matrix of the neural object computed above, we can simulate its behavior in free space. However, contact interactions with the environment or with a robot require us to compute normal and friction forces. In physics simulation, the contact force direction and magnitude are computed to limit interpenetration between the object and its environment \cite{lee2018dart, elandt2019pressure}. This requires quantifying the location and the amount of interpenetration between an object and its environment. 
        Classical formulations of contact interaction rely on signed distance functions (SDFs) \cite{heiden2021neuralsim} to measure this quantity. With neural objects, we only have access to the volumetric density field $\phi: \R^3 \rightarrow \R_{+}$, which has no distinct notion of a surface.
    
        The core idea behind our contact model is to measure the amount of interpenetration between two objects. Given two objects $A$ and $B$, we measure the overlap as the integral,
        \begin{equation}
            \psi = \int_{x \in \R^3} \phi^A (x) \phi^B(x) dx,
            \label{eq:integral_volume}
        \end{equation}
        of the product of their density fields $\phi^A$ and $\phi^B$ over the whole workspace $\R^3$.
        A probabilistic interpretation of this expression gives an intuitive justification for this choice. Let us assume that $\phi^A(x) = P(H^A_x = 1)$ is the probability of hitting object $A$ when sampling a point $x$ in $\R^3$.
        \footnote{In practice, $\phi$ is an unnormalized density field; nevertheless, our contact model only requires computing volume $\psi$ up to a scaling factor.}. 
        Here, $H^A_x$ is a random variable taking value $1$ to indicate collision and $0$ otherwise. Then assuming $H^A_x$ and $H^B_x$ are independent random variables (i.e., the shape of one object is independent of the shape of the other), $\phi^A(x) \cdot \phi^B(x) = P(H^A_x = 1 \cap H^B_x = 1)$ is the probability of hitting both objects $A$ and $B$ when sampling point $x$. 
        Finally, $\psi$ (Eq.~\ref{eq:integral_volume}) represents the \emph{expected interpenetration volume} between objects $A$ and $B$. We let the amount of repulsive force applied between objects $A$ and $B$ be proportional to $\psi$.
        
        Additionally, we compute the \emph{expected centroid of the interpenetration volume}. This is the geometric center of the interpenetration volume and the point at which we apply the repulsive contact forces,
        \begin{equation}
            \chi = \frac{1}{\psi} \int_{x \in \R^3} x  \phi^A (x) \phi^B(x) dx, \quad \text{if} \: \: \psi \neq 0.
            \label{eq:integral_centroid}
        \end{equation}
        When $\psi = 0$, there is no interpenetration, and the computation of $\chi$ is unnecessary. 
        
        \textbf{Neural Object Sampling Procedure.} Exact computation of the integrals in Eq.~(\ref{eq:integral_volume}, \ref{eq:integral_centroid}) are intractable. 
        We approximate them using a Monte-Carlo sampling scheme. As with the mass, center of mass, and inertia matrix computations, we sample a set of points $X$ uniformly from the neural density field $\phi(x)$ workspace. For contact interactions, we keep the points between a minimal and maximal density value and discard the rest, since low-density points tend to be outside the object, and high-density points tend to be farther in the interior of the object. This biases the sampling of points towards the boundary of the object. This way, we densely cover the boundary of the object while limiting the number of sample points for computational efficiency. 
        In all the experiments presented in this paper, we use $5000$ sample points. 
        We denote $X^A$ the set of $N^A$ points sampled from object $A$.
  
    \begin{figure}[t]
    	\begin{center}
    	\includegraphics[width=0.6\linewidth]{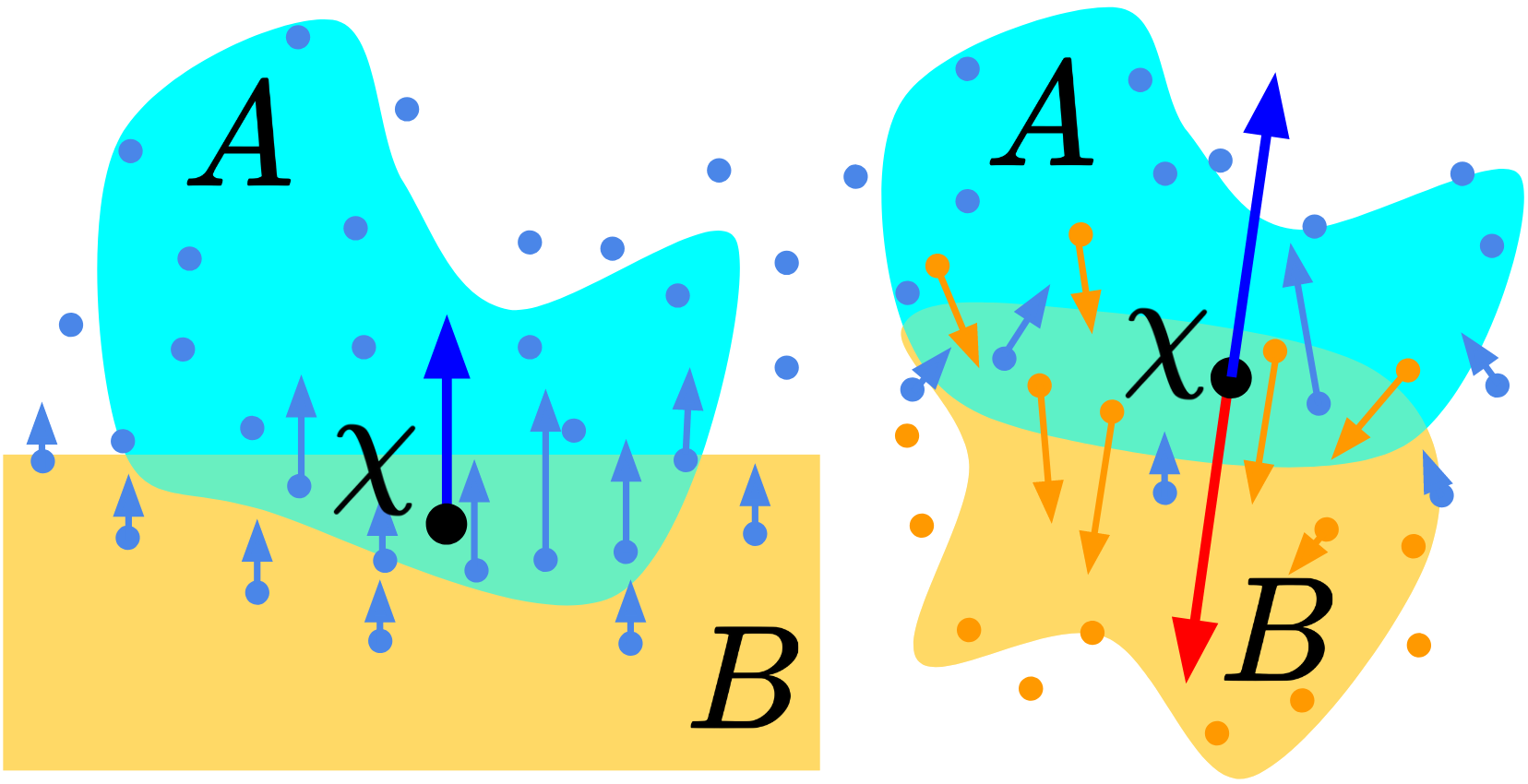}
    	\end{center}
        \vspace{-3mm}
        \caption{Left: contact between a DANO ($A$) and the ground (half-space $B$). Each point sampled on the density field below ground level contributes to the repulsive force on the DANO. Points with a larger density value generate more force. The repulsive forces are applied at point $\chi$, which represents the geometric center of the interpenetration volume between object $A$ and $B$. The direction of the repulsive force is the outward normal of the half-space.  
        Right: contact between two DANOs. Points, where both objects have large density values, generate large repulsive forces. Each repulsive force is directed along its local density field gradient and contributes to the overall contact normal proportionally to its magnitude.}
    	\label{fig:contact_normal}
        \vspace{-3mm}
    \end{figure}

    \textbf{Contact Between Neural Object \& Shape Primitive.}
        To illustrate our contact model, we choose a simple scenario (Fig. \ref{fig:contact_normal}, left) where a neural object $A$ collides with the ground represented by a half-space $B$.  The shape primitive can be expressed as $\mathcal{B} = \{x \in \R^3 | f(x) \leq 0\}$ where, for example, $f(x) = a^Tx + b$ for a half-space, or $f(x) = || x - c || - r$ for a sphere with center $c$ and radius $r$. We represent the density function for the shape primitive as $\phi^B(x) = \mathbf{1}_{x \in \mathcal{B}}$, where $\mathbf{1}$ denotes the indicator function. 
        Finally, we approximate the interpenetration volume and its centroid Eq.~(\ref{eq:integral_volume},  \ref{eq:integral_centroid}) as sums over sampled points:
            \begin{align}
            \vspace{-3mm}
                \psi &\approx \frac{1}{N} \sum_{x \in X^A} \phi^A(x) \phi^B(x)
                    = \frac{1}{N} \sum_{x \in X^A \cap \mathcal{B}} \phi^A(x), \\
                \chi &\approx \frac{1}{\psi N} \sum_{x \in X^A \cap \mathcal{B}} x \phi^A(x).
            \end{align}            
        Each sampling point $x \in X^A$ is expressed in a frame attached to object $A$. Thus the value of $\phi^A(x)$ is independent of the position and orientation of objects $A$ and can be computed offline. Translations and rotations of object $A$ influence the values of $\psi$ and $\chi$, modifying the number of sampling points belonging to $\mathcal{B}$. For example, if object $A$ is positioned halfway through the ground, $\psi$ will be large, whereas if object $A$ is above the ground, $\psi$ should be close to zero.
        Since $\phi^A(x)$ is pre-computed, we can simulate contact without resampling the density field online. The contact normal $n$ is the outward facing normal to the shape primitive computed at the point $\chi$, e.g. $n = (\chi - c) / ||\chi - c||_2$ for a sphere centered in $c$.
        We implement this to simulate contact with half-spaces and spheres, but this approach generalizes to a variety of shape primitives, including capsules, boxes, and compositions of shape primitives. 
    
    \textbf{Contact Between Two Neural Objects.}
        We follow the same approach to model contact between two neural objects $A$ and $B$ (Fig. \ref{fig:contact_normal} right). We process object $A$'s density field to extract a set of $N^A$ sampled points $X^A$ expressed in a frame attached to object $A$. We do the same for object $B$. Then our Monte-Carlo sampling scheme gives
        \begin{align}
            \vspace{-3mm}
            \psi &\approx \frac{1}{N^A + N^B} \sum_{x \in X^A \cup X^B} \phi^A(x) \phi^B(x), \\
            \chi &\approx \frac{1}{\psi (N^A +N^B)} \sum_{x \in X^A \cup X^B} x \phi^A(x) \phi^B(x).
        \end{align}
        We precompute $\phi^A(x^A)$ for all $x^A$ in $X^A$ and $\phi^B(x^B)$ for all $x^B$ in $X^B$. However, the value of $\phi^A(x^B)$ for $x^B$ in $X^B$ varies with the relative configuration of objects $A$ and $B$. Thus, we compute these quantities online each time the relative position of the two objects changes. Identifying the normal to the contact requires both offline and online computing. 
        Offline, we sample outward facing normals aligned with the neural density field gradient $n(x) = - \nabla_x \phi(x) / ||\nabla_x \phi(x)||_2$ using a finite-difference scheme (blue arrows Fig~\ref{fig:preprocess}). Online, we compute the normal $n$ to the contact as a weighted average of the offline sampled contact normals,
        \begin{align*}
            \bar{n} &\approx \sum_{x \in X^A} n^A(x) \phi^A(x) \phi^B(x)
            - \sum_{x \in X^B} n^B(x) \phi^A(x) \phi^B(x)
        \end{align*}
        where $n = \bar{n} / ||\bar{n}||_2$. The minus sign accommodates for the opposite direction of $n^A(x)$ and $n^B(x)$ as illustrated in Fig~\ref{fig:contact_normal}.
        
        \begin{table}[t]
    	\centering
    	\caption{Contact model parameters. For each parameter, we provide a nominal range of values that produce realistic physical behavior. Additionally, we describe how changing this value affects the simulation.}
    	\begin{tabular}{c c c c}
    		\toprule
    		\textbf{parameter} & \textbf{min} & \textbf{max} & \textbf{effect}\\
    		\toprule
    		impact spring & $10^{4}$ & $10^5$ & $\nearrow$ stiffer impact\\
    		impact damper & $10^{5}$ & $10^6$ & $\nearrow$ more stable simulation\\
    		\hline
    		sliding friction & 0 & 1 & $\nearrow$ less sliding\\
    		sliding drag & 0 & 0.1 & $\nearrow$ more stable simulation\\
    		\hline
    		rolling friction & 0 & 0.1 & $\nearrow$ less rolling\\
    		rolling drag & 0 & 0.1 & $\nearrow$ more stable simulation\\
    		\toprule
    		torsional friction & 0 & 0.1 & $\nearrow$ less spinning\\
    		torsional drag & 0 & 0.1 & $\nearrow$ more stable simulation\\
    		\toprule
    	\end{tabular}
         \vspace{-3mm}
    	\label{tab:contact_model}
    \end{table}

    \begin{figure}[t]
        \begin{center}
        \includegraphics[width=0.50\columnwidth]{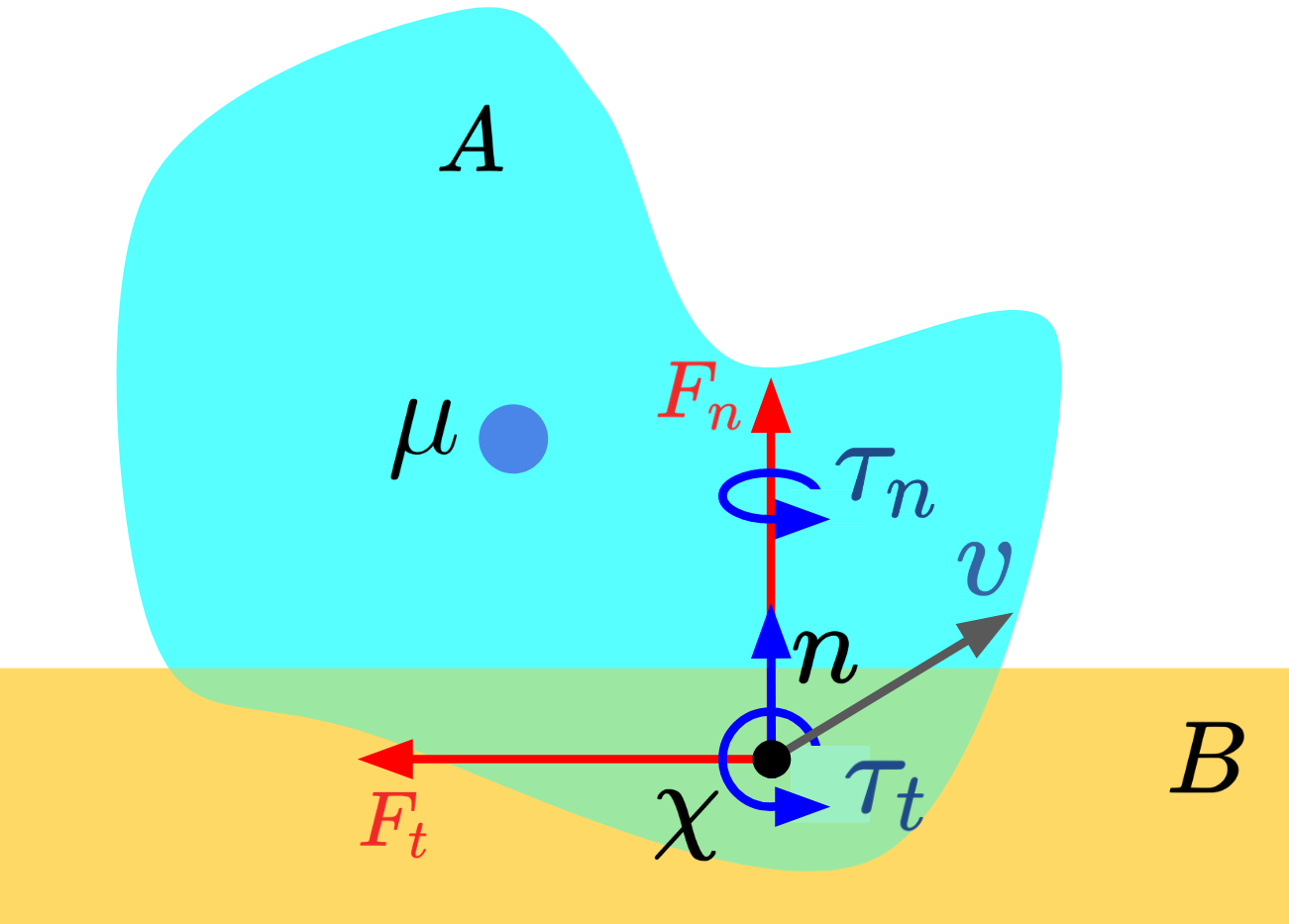}
        \caption{
        Contact modeling between a DANO $A$ and a primitive shape $B$ (half-space). All forces and torques are applied at the centroid of the interpenetration volume $\chi$. The normal force $F_n$ opposing contact interpenetration is applied along the contact normal $n$. A tangential force $F_t$ modeling sliding friction is applied. This force opposes the velocity of the point $\chi$ in the plane tangential to the contact. Similarly, rolling friction is applied via a torque $\tau_t$ in the tangential plane. Finally, $\tau_n$ generates torsional friction.
        } \label{fig:contact_model}
        \end{center}
        \vspace{-3mm}
    \end{figure}

    \textbf{Computing Contact Forces.} We compute contact forces and torques proportional to the interpenetration volume $\psi$, applied at the centroid of the interpenetration volume $\chi$, and applied in the direction normal to the contact surface, as shown in Fig.~\ref{fig:contact_model}). 
    We define the linear velocity $v \in \mathbf{R}^3$ of point $\chi$ attached to frame $A$ with respect to frame $B$, and the angular velocity $\omega \in \mathbf{R}^3$ of frame $A$ with respect to frame $B$.
    We apply a force normal to the contact to oppose interpenetration,
    \begin{align}
        F_n = -\psi \left( I_{\mbox{spring}} n + I_{\mbox{damper}} v_n \right),
    \end{align}
    where $n$ is the unit vector normal to the contact, and $v_n$ is the component of the velocity $v$ normal to the contact. 
    
\begin{figure}[t]
		\begin{center}
            \begin{tikzpicture}
                \draw (0, 0) node[inner sep=0] {\includegraphics[width=\linewidth]{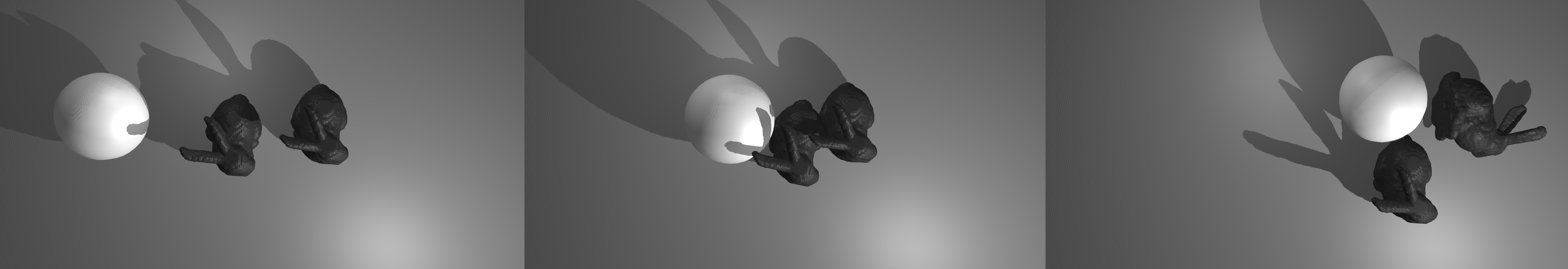}};
                \draw (-3.0, -0.60) node {\textcolor{white}{$t = 0.0 s$}};
                \draw (+0.0, -0.60) node {\textcolor{white}{$t = 0.2 s$}};
                \draw (+3.0, -0.60) node {\textcolor{white}{$t = 1.0 s$}};
            \end{tikzpicture}
		\end{center}
        \vspace{-3mm}
		\caption{We simulate an environment with a sphere hitting two neural objects (Stanford bunnies). During simulation, the two neural objects make and break contact propagating the impulse provided by the hitting sphere.}
        \vspace{-3mm}
		\label{fig:twoDANOs}
	\end{figure}
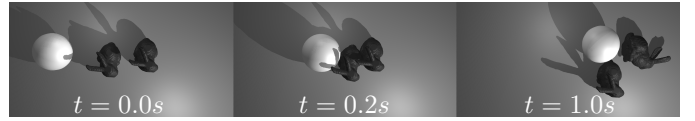

    $I_{\mbox{spring}}$ and $I_{\mbox{damper}}$ are parameters modeling the stiffness of the contact. 
    We apply a force tangential to the contact to oppose relative sliding between objects $A$ and $B$
    \begin{align}
        F_t = -||F_n|| \left( S_{\mbox{friction}} \frac{v_t}{||v_t||} + S_{\mbox{drag}} v_t \right),
    \end{align}
    where $v_t$ is the tangential component of $v$, $S_{\mbox{friction}}$ and $S_{\mbox{drag}}$ are parameters modeling dry and viscous frictions. 
    We apply a torque normal to the contact to encode torsional friction,
    \begin{align}
        \tau_n = -||F_n|| \left(T_{\mbox{friction}} \frac{\omega_n}{||\omega_n||} + T_{\mbox{drag}} \omega_n \right),
    \end{align}
    where $\omega_n$ is the normal component of $\omega$, $T_{\mbox{friction}}$ and $T_{\mbox{drag}}$ are parameters modeling dry and viscous torsional frictions. 
    Finally, we apply rolling friction that opposes the relative rotation of object $A$ with respect to object $B$ when they are in contact,
    \begin{align}
        \tau_t = -||F_n|| \left(R_{\mbox{friction}} \frac{\omega_t}{||\omega_t||} + R_{\mbox{drag}} \omega_t \right),
    \end{align}
    where $\omega_t$ is the tangential component of $\omega$, $T_{\mbox{friction}}$ and $R_{\mbox{drag}}$ are parameters modeling dry and viscous rotational frictions. 
    In Table \ref{tab:contact_model}, we provide default values and an explanation of the effect of each parameter on contact simulation. 
    We acknowledge that the simple contact model proposed in this paper is subject to creep, (e.g. an object slowly slides on an inclined plane instead of sticking). 
    This is a limitation commonly seen in many existing physics engines \cite{todorov2012mujoco}. To address this issue, an optimization-based modeling of friction forces and torques, as proposed in \cite{howelllecleach2022}, could be implemented.
    To include contact normal and tangential forces and torques in the simulator, we compute $F$ the contact force applied by body $B$ on body $A$ expressed in the world frame, $\tau$ the torque evaluated at the center of mass $\mu$ of body $A$ expressed in body $A$'s frame, 
    \begin{align}
        \vspace{-6mm}
        F &= F_n + F_t, \\
        \tau &= \tau_n + \tau_t + \overrightarrow{\mu \chi} \times F.
    \end{align}
    We use a first-order variational integrator identical to Dojo's~\cite{howelllecleach2022}. In addition, forces related to neural object contacts are computed explicitly at the current configuration and integrated over one time step.
    To illustrate the DANO and contact force model described above in a complex simulation scenario, we show a simulation run with contact interactions between two Stanford bunny DANOs, a sphere, and a half-space (ground) in Fig~\ref{fig:twoDANOs}. Please refer to the supplemental video \href{https://youtu.be/Md0PM-wv_Xg}{$\texttt{youtu.be/Md0PM-wv\_Xg}$} for animations of multiple different simulation scenarios. An implementation in the Julia language of the method and applications is publicly available \href{https://github.com/dojo-sim/Dojo.jl/tree/DANO}{$\texttt{github.com/dojo-sim/Dojo.jl/tree/DANO}$}.
    
    \section{Applications}
    \label{sec:applications}
        In this section, we use a DANO model of a bar of soap acquired from real images of the soap.  We find the coefficient of friction of the soap from a real video of the soap sliding on a table. We then use a DANO model of the Stanford bunny (acquired from synthetic images of the bunny) to find the friction coefficient and mass scaling coefficient of the bunny (in simulation).  Finally, we optimize a robot trajectory in simulation for pushing the bunny to a goal location using the differentiability of the simulation pipeline. 
    
    \begin{figure}[t]
        \centering
      \subfloat[
            Left: we visualize how closely the dynamics-augmented neural object with learned friction matches the ground-truth trajectory. Top: simulated trajectory poorly matching the ground truth using an initial guess for the sliding friction coefficient. Middle: simulated trajectory closely matching the ground-truth trajectory with learned sliding friction coefficients. Bottom: ground-truth trajectory extracted from a video of the soap sliding on the ground. Right: the trajectory prediction error (Problem \ref{pb:system_identification}) rapidly decreases after a few Newton steps.
      \label{fig:system_identification_soap}
      ]{%

        \begin{tikzpicture}
            \draw (0, 0) node[inner sep=0] {\includegraphics[height=22mm]{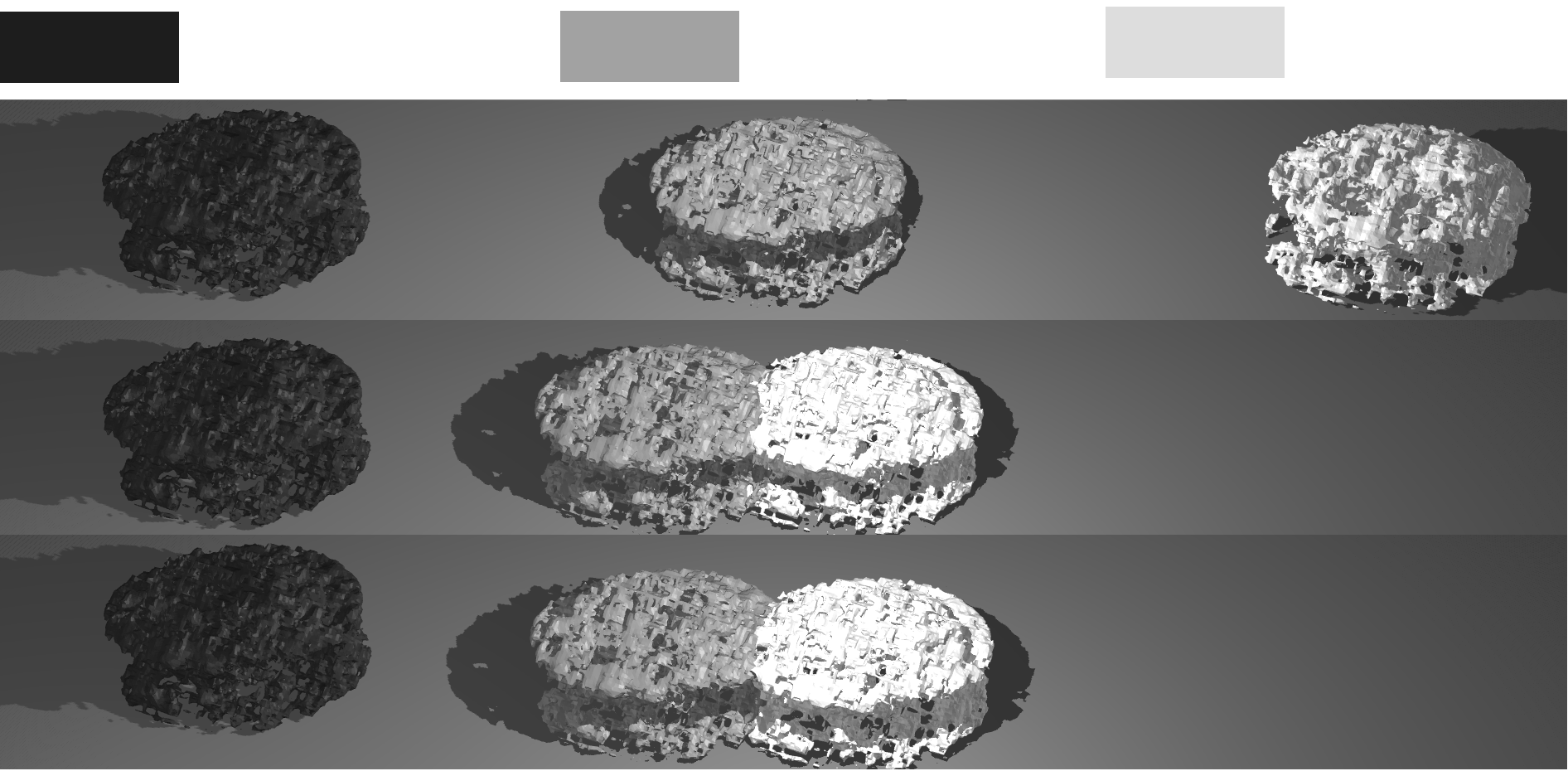}};
            \draw (-2.7, +0.65) node {\footnotesize{initial}};
            \draw (-2.7, +0.35) node {\footnotesize{guess}};
            \draw (-2.7, -0.1) node {\footnotesize{learned}};
            \draw (-2.7, -0.65) node {\footnotesize{ground}};
            \draw (-2.7, -0.95) node {\footnotesize{truth}};
            \draw (-1.7, +1.2) node {\footnotesize{$t = 0.0 s$}};
            \draw (-0.05, +1.2) node {\footnotesize{$t = 0.14 s$}};
            \draw (+1.45, +1.2) node {\footnotesize{$t = 0.28 s$}};
        \end{tikzpicture}
		\includegraphics[height=28mm]{tikz/bluesoap_learning.tikz}}
        \hfill
        
      \subfloat[Left: we learn the mass, inertia scaling, and sliding friction coefficient of the bunny through interactive perception. We use a spherical end effector to push a neural object (bunny). This interaction facilitates the identification of dynamics parameters such as mass. Right: we closely match ground-truth parameters and trajectories using a dataset of 10 pushes.
        \label{fig:system_identification_bunny}
        ]{%
        \begin{tikzpicture}
            \draw (0, 0) node[inner sep=0] {\includegraphics[height=25mm]{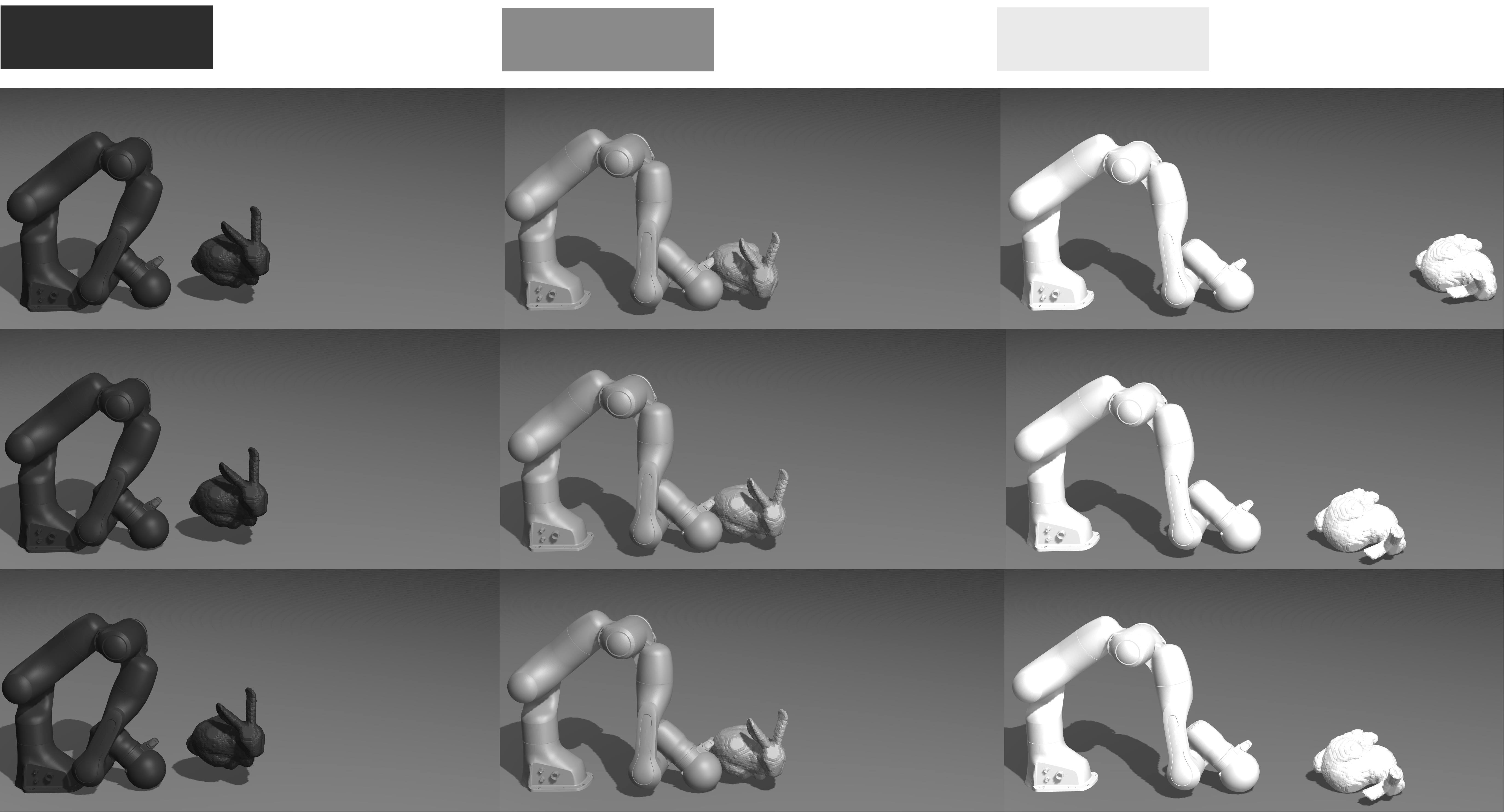}};
            \draw (-2.8, +0.75) node {\footnotesize{initial}};
            \draw (-2.8, +0.45) node {\footnotesize{guess}};
            \draw (-2.8, -0.2) node {\footnotesize{learned}};
            \draw (-2.8, -0.75) node {\footnotesize{ground}};
            \draw (-2.8, -1.05) node {\footnotesize{truth}};
            \draw (-1.8, +1.4) node {\footnotesize{$t = 0.0 s$}};
            \draw (-0.3, +1.4) node {\footnotesize{$t = 0.2 s$}};
            \draw (+1.2, +1.4) node {\footnotesize{$t = 2.0 s$}};
        \end{tikzpicture}
        \includegraphics[height=28mm]{tikz/sphere_bunny_learning.tikz}}            
      \caption{Results on system identification.}
      \vspace{-3mm}
      \label{fig:system_identification} 
    \end{figure}

    \subsection{System Identification}
    \label{sec:SysID}
        We demonstrate how we can leverage differentiable simulation to efficiently estimate the dynamical properties of the DANO from object trajectories.  
        For a given object, we parameterize our dynamics model with a vector $\theta \in \R^d$. This vector can include contact-model parameters, such as sliding or rolling-friction coefficients, and dynamics parameters, such as the mass and inertia of the object. 
        We learn $\theta$ by minimizing the distance between a ground-truth trajectory and a simulated trajectory starting from the same initial conditions. 
        \begin{equation}
        	\begin{array}{ll}
        	\underset{\theta}{\mbox{minimize}} & \sum_{t=2}^{T} \| \hat{x}_t - x_t \|_W^2\\
        	\mbox{subject to} & x_{t+1} = f(x_t; \theta), \quad t = 1, \dots, T-1,\\
        	& x_1 = \hat{x}_1,\\
        	& \theta_{\text{min}} \leq \theta \leq \theta_{\text{max}},\\
        	\end{array} \label{pb:system_identification}
        \end{equation}
        where we indicate ground-truth quantities with the $\,\hat{}\,$ symbol, $x_1$ is the system's initial condition, $\| \cdot \|_W$ is a weighted norm, $f$ is the dynamics parameterized by $\theta$, and $T$ is the number of time steps. $\theta_{\text{min}}$ and $\theta_{\text{max}}$ are bounds on the parameters enforcing basic constraints; for example, mass and friction coefficients necessitate positive quantities. 
        Leveraging the simulator's differentiability i.e., $f$'s derivatives, we apply the Gauss-Newton method \cite{Nocedal2006} to learn the system's parameters.
        
        \textbf{Soap Bar.} We apply system identification on real-world data (Fig~\ref{fig:system_identification_soap}). We train an OSF model from a set of still images of a semi-translucent soap bar.  Then we extract a pose trajectory from a video of the soap bar sliding on the ground.  We formulate the pose tracking as a frame-wise optimization problem using only geometric information:
            $\min_{p} \lVert T(p)- M\rVert^2 + \lVert B(T(p)) - B(M)\rVert^2$,
        where $p$ denotes the object pose at the current frame, $M$ denotes a binary object mask extracted from the frame (we use the U$^2$Net~\cite{qin2020u2}), $T(p)$ denotes the accumulated transmittance map rendered using object pose $p$, and $B(\cdot)$ denotes the barycenter of the mask/transmittance map. The accumulated transmittance map is computed along with the volume rendering process~\cite{mildenhall2020nerf}, and the barycenter can be easily computed by the weighted mean of the mask/transmittance map multiplied by pixel coordinates. We solve this optimization problem using the Adam optimizer for all video frames to obtain the pose trajectory.
        
        We learn the sliding friction coefficient between the soap and the table from this trajectory. The optimization process takes 2.0 seconds on a laptop equipped with an Intel i7-8750H CPU and 16GB of RAM. The resulting trajectory simulated with learned friction closely matches the ground-truth trajectory. To find the ground truth coefficient of friction, we experimentally collected trajectory data where the bar of soap is sliding on a tilted plane with a known angle. We obtained a ground-truth value of $S_{\mbox{friction}} = 0.75$ which is close to the value obtained from system identification: $S_{\mbox{friction}} = 0.61$.
    
        \textbf{Stanford Bunny.} In simulation, we leverage interactive perception~\cite{bohg2017interactive} to identify the mass scaling factor $\alpha$ from (\ref{eq:MassScale}), and sliding friction coefficient between a dynamics-augmented neural object and a surface (Fig.~\ref{fig:system_identification_bunny}). 
        We implement a simple policy where the end effector pushes the DANO bunny with a known force, thereby allowing the identification of both the mass scale factor and friction coefficient. We successfully identify these parameters with less than $1.5\%$ relative error from 10 pushing trajectories.
    
    \subsection{Trajectory Optimization}
        Simulating the DANO in a differentiable simulator such as Dojo \cite{howelllecleach2022} allows us to optimize robot trajectories that involve contact (e.g., grasping, manipulation, or pushing) using existing gradient-based optimization frameworks. For trajectory optimization, we minimize a cost functional over a time-discretized trajectory of the robot while respecting state and control-input constraints,
        \begin{equation}
        	\begin{array}{ll}
        	\underset{x_{1:T}, u_{1:T-1}}{\mbox{minimize }} & \sum_{t=1}^{T-1} l_t(x_t, u_t) + l_T(x_T) \\
        	\mbox{subject to} & x_{t+1} = f_t(x_t, u_t), \quad t = 1, \dots, T-1,\\
        	& x_1 = \hat{x}_1,\\
        	& c_t(x_t, u_t) \leq 0, \quad \phantom{\,\,\,\,\,\,\,\,\,} t = 1, \dots, T-1,\\
        	& c_T(x_T) \leq 0,\\
        	\end{array} \label{pb:trajectory_optimization}
        \end{equation}
        where the subscript $t$ indicates the time step, $T$ is the number of time steps, $x$ is the system state, $u$ is the control input, $\hat{x}_1$ is the system's initial conditions, $l_t$ and $l_T$ are stage and final cost functions respectively, $c_t$ and $c_T$ are stage and final constraints respectively, and $f_t$ is the dynamics.
        
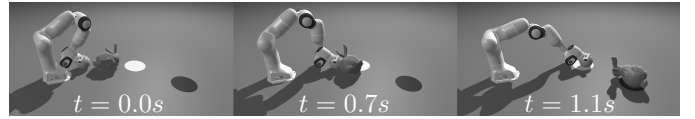
\begin{figure}[t]
    		\begin{center}
                \begin{tikzpicture}
                    \draw (0, 0) node[inner sep=0] {\includegraphics[width=\linewidth]{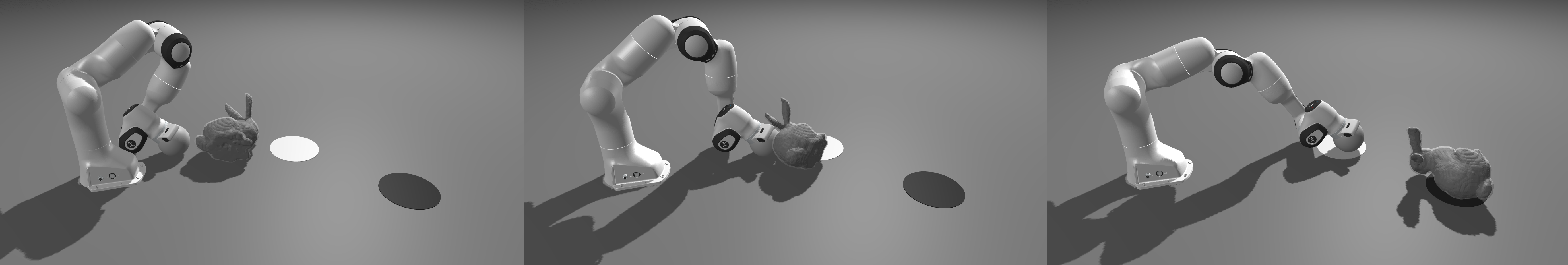}};
                    \draw (-3.0, -0.60) node {\textcolor{white}{$t = 0.0 s$}};
                    \draw (+0.0, -0.60) node {\textcolor{white}{$t = 0.7 s$}};
                    \draw (+3.0, -0.60) node {\textcolor{white}{$t = 1.1 s$}};
                \end{tikzpicture}
    		\end{center}
            \vspace{-3mm}
    		\caption{We solve a push-and-slide task using trajectory optimization. The objective is to push a dynamics-augmented neural object (bunny) using a fully-actuated spherical end effector (Panda arm). The goal positions of the end effector and the bunny are shown with white and black circular targets, respectively. Leveraging the simulator's differentiability, we optimize with a gradient-based solver a highly dynamic trajectory where the bunny slides to reach its target position.
    		}
            \vspace{-3mm}
    		\label{fig:trajectory_optimization}
    	\end{figure}

        We optimize dynamic behaviors with contact on a push-and-slide task (Fig.~\ref{fig:trajectory_optimization}). Specifically, a simulated Panda robot arm tries to push a DANO model of the Stanford bunny to a goal location, and return the Panda's end effector to another goal location. We use a constrained iterative linear quadratic regulator (iLQR) solver \cite{li2004iterative, howell2019altro}, which exploits gradients of the simulation.
    
    \section{Comparison with Mesh-based Simulation}
    \label{sec:mesh}
        It is natural to ask whether our DANO approach offers benefits over an approach leveraging existing methods and tools. A reasonable approach could be to convert the neural object representation into a mesh and pass this representation to an existing differentiable simulator. In this section, we investigate this approach to point out several unforeseen difficulties emerging with mesh-based simulation methods.
    
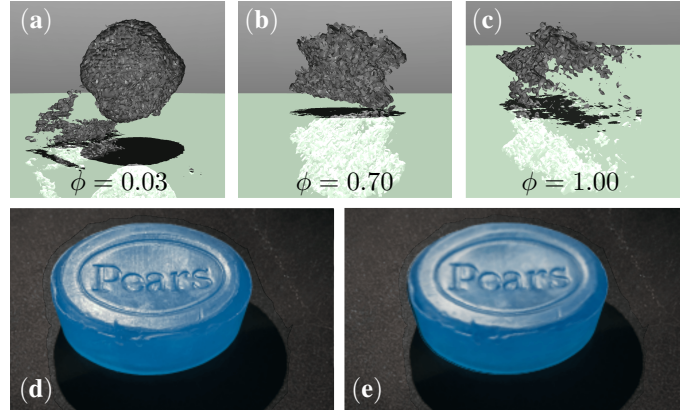
\begin{figure}[t]
    		\begin{center}
                \begin{tikzpicture}
                    \draw (0, 0) node[inner sep=0] {\includegraphics[width=1.0\linewidth]{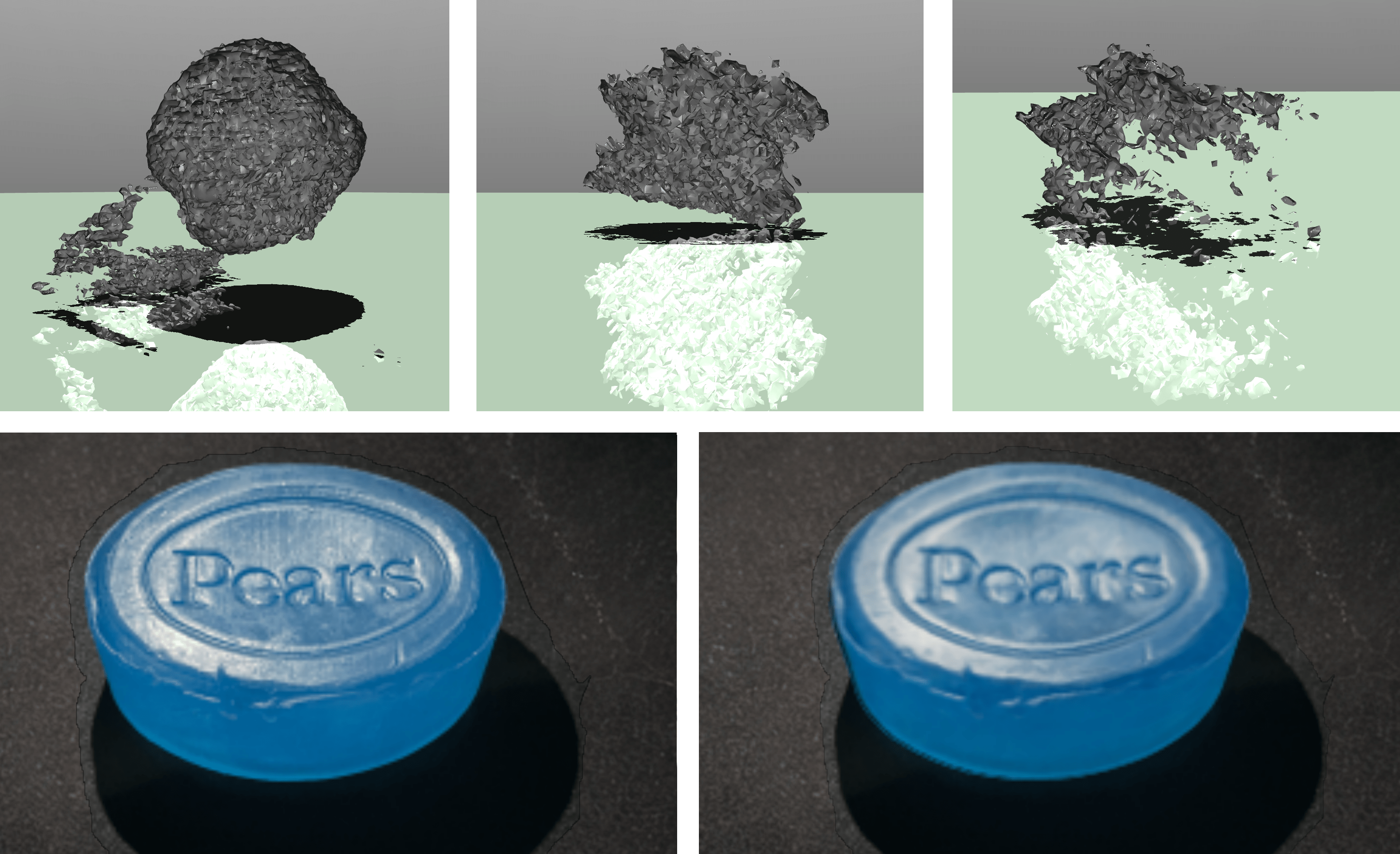}};
                    \draw (-3.0, +1.17*0.25) node {$\phi = 0.03$};
                    \draw (+0.0, +1.17*0.25) node {$\phi = 0.70$};
                    \draw (+3.0, +1.17*0.25) node {$\phi = 1.00$};
                    \draw (-4.1, -1.17*2.1) node {\textcolor{white}{$(\textbf{d})$}};
                    \draw (+0.3, -1.17*2.1) node {\textcolor{white}{$(\textbf{e})$}};
                    \draw (-4.1, +1.17*2.1) node {\textcolor{white}{$(\textbf{a})$}};
                    \draw (-1.1, +1.17*2.1) node {\textcolor{white}{$(\textbf{b})$}};
                    \draw (+1.9, +1.17*2.1) node {\textcolor{white}{$(\textbf{c})$}};
                \end{tikzpicture}
    		\end{center}
            \vspace{-3mm}
    		\caption{\textbf{(a, b, c)} Meshes extracted from 3 different level set values of the neural density, simulated lying on the floor at rest in MuJoCo. 
            \textbf{(a)} With a low level-set value of $\phi = 0.03$, spurious mesh artifacts float around the actual soap bar, preventing it from actually contacting the floor. \textbf{(b)} The level-set value $\phi = 0.70$ best captures the original shape of the soap bar, but there are still spurious artifacts that cause the soap to lie on the ground with an unnatural tilt.  \textbf{(c)} The level-set $\phi = 1.00$ allows the soap to lie flat on the floor; however, the object is mostly void, leading to inaccurate mass, center of mass, and inertia matrix computations. Simulations with contact will be highly unreliable when most of the object is modeled as free space.
            \textbf{(d)} Picture of the soap bar, \textbf{(e)} Simulation of the DANO at rest on the floor, it lies flat safely ignoring spurious artefacts. 
            }
    		\label{fig:mesh_simulation}
            \vspace{-6mm}
    	\end{figure}

        \textbf{Level-set Selection.}
        To extract a mesh from a neural density field, we apply the Marching Cubes algorithm \cite{lorensen1987marching, newman2006survey} on the underlying OSF density field. This extraction requires picking a single density level set from the OSF, which begs the question, which density value should we choose? To answer this question, we extract a series of 3 meshes from an OSF density field of a bar of soap (constructed from real images). Similarly to DANO, we augment these level-set objects with dynamical properties (e.g. mass, inertia, friction). Then we simulate these level-set objects in an existing physics engine: MuJoCo \cite{todorov2012mujoco}. Simulation results and visualizations of the level-set objects are presented in Figure~\ref{fig:mesh_simulation}.  Finding a suitable density value is a labor-intensive task, as visual inspection of the generated mesh is often not sufficient due to small masses floating around the object.

        \textbf{Mesh Artifacts.} 
        Once a density value is chosen, the resulting level-set object often features mesh artifacts preventing accurate contact simulation. For lower-density values (Fig. \ref{fig:mesh_simulation}(a)), contact interaction with the ground is inaccurate due to masses floating around the object. Small artifacts persist even for larger-density values (Fig. \ref{fig:mesh_simulation}(b)) and dictate the contact interactions.       
        In contrast, DANO's contact model is less sensitive to such artifacts. Indeed, small volumes with high densities floating around the object will contribute little to the contact interaction (Eq. \ref{eq:integral_volume}).
        For higher-density values (Fig. \ref{fig:mesh_simulation}(c)), the object is mostly void, and contact interaction with robotic hardware such as a 2-finger gripper cannot be accurately simulated.  To the best of our knowledge, there is no standard algorithm for fixing either of these two types of artifacts. Fixing these issues might require manual intervention or labor-intensive tuning of an ad-hoc method. Figure~\ref{fig:mesh_simulation}(d) shows an image of the true object and Fig.~\ref{fig:mesh_simulation}(e) shows the pose of our DANO lying flatly on the ground at rest, but rendered with a mesh for easy visualization.  Please note the poor quality of the mesh in Fig.~\ref{fig:mesh_simulation}(d) has nothing to do with the physics simulation, and is just for visualization. 
    
        \textbf{Contact Simulation.}
        Finally, existing physics engines targeting robotics applications have limited support for mesh-represented objects. Among the major simulators: MuJoCo \cite{todorov2012mujoco}, Bullet \cite{heiden2021neuralsim}; none can directly simulate non-convex mesh-represented objects\footnote{For instance, MuJoCo approximates non-convex meshes with a convex hull.}. A pre-processing step decomposing the object into a set of convex shapes is required. Voxelized Hierarchical Approximate Convex Decomposition (V-HACD) \cite{mamou2016volumetric} is a commonly used algorithm with its own set of tuning parameters; prominently the number of convex shapes used to represent the object.\footnote{
        For completeness, we mention that we applied a convex-decomposition to each level-set mesh~\cite{wei2022approximate}. However, due to the chaotic surface of the meshes, the resulting decomposition featured sharp and tiny convex shapes that were not handled by the MuJoCo simulator.
        } Comparatively, DANO's approach has been demonstrated with non-convex objects like the Stanford bunny (Figure \ref{fig:system_identification} \& \ref{fig:trajectory_optimization}) without requiring additional convex decomposition, manual tuning, or multiple algorithmic stages with human oversight and input.
        
        Overall, the traditional mesh-based simulation approach requires, in practice, multiple mesh processing steps and labor-intensive parameter tuning involving several arbitrary choices (density value selection, convex decomposition). Our DANO approach bypasses mesh-related issues by directly operating on the density field.
               
    \section{Discussions} We presented the Dynamically-Augmented Neural Object (DANO), which appends a neural object model with essential dynamical information, making its motion simulatable in a physics simulation environment. We also propose a contact force model to compute normal and friction forces for the DANO. We see these tools as an effort to bridge the gap between perception and simulation in robotics. Ultimately, we hope this work is a step towards endowing a robot with the ability to autonomously build its own physics simulations of its environment using only its own sensor inputs---an essential ingredient of robot spatial intelligence. 

    \textbf{Limitations.} Our method does have some noteworthy limitations.
    (i) One limitation of our method is that out current implementation applies forces at a single point. One could instead consider all the sampled points independently to better handle torsional friction and strongly non-convex objects. 
    (ii) Additionally, articulated body and soft body contact simulation is an interesting and important topic for future research. These kinds of objects are not covered in our current work, as rigid body objects present many challenges on their own.  
    (iii) Another limitation of our method is that extracting rigid body pose trajectories from RGB video is challenging. Our pose extraction method in Sec.~\ref{sec:SysID} uses a 2D mask-based loss, which has numerous local minima for object orientation and thus is sensitive to initialization. A better approach would be to solve for the pose trajectory by minimizing the photometric error between the actual video and a video generated by the neural object renderer and the differentiable simulator together. (iv) This points to the second limitation, which is that the method in this paper stops one step short of integrating the dynamics simulator with the renderer to give a differentiable end-to-end ``torques-to-pixels'' simulator. This is possible with the tools we describe, and we plan to pursue this as our immediate next step.  Finally, (v) in real videos, the apparent color of a point on the object changes as it moves relative to the ambient light field, changing reflections, specularity, shadows, and color. We ignore these effects in this paper, although it is possible to reproduce such effects with the OSF model. It would be interesting to attempt to capture these lighting effects in an end-to-end differentiable ``torques-to-pixels'' simulator.

\bibliographystyle{ieeetr} 
\bibliography{reference}  

\end{document}